%% file: camera_version.tex
\documentclass{bmvc2k}

\usepackage{epsfig}
\usepackage{graphicx}
\usepackage{amsmath}
\usepackage{amssymb}
\usepackage{soul}
\usepackage{caption,booktabs,multirow,comment}
\usepackage{enumitem}
\usepackage{array}
\newcolumntype{L}[1]{>{\raggedright\let\newline\\\arraybackslash\hspace{0pt}}m{#1}}
\newcolumntype{C}[1]{>{\centering\let\newline\\\arraybackslash\hspace{0pt}}m{#1}}
\newcolumntype{R}[1]{>{\raggedleft\let\newline\\\arraybackslash\hspace{0pt}}m{#1}}

\newcommand{\specialcell}[2][c]{%
  \begin{tabular}[#1]{@{}c@{}}#2\end{tabular}}
  
\newcommand{\etal}{{\em et al. }}
\definecolor{blue}{rgb}{0,0,.4}  
\usepackage[font={small, color=blue},skip=5pt]{caption}
\usepackage{pifont}
\usepackage{xpatch}
\makeatletter
\xpatchcmd{\paragraph}{3.25ex \@plus1ex \@minus.2ex}{3pt plus 1pt minus 1pt}{\typeout{success!}}{\typeout{failure!}}
\makeatother
\bmvcreviewcopy{669}

\title{A Closer Look at Temporal Ordering in the Segmentation of Instructional Videos}

\addauthor{Anil Batra}{A.K.Batra@sms.ed.ac.uk}{}
\addauthor{Shreyank N Gowda}{S.Narayana-Gowda@sms.ed.ac.uk}{}
\addauthor{Frank Keller}{keller@inf.ed.ac.uk}{}
\addauthor{Laura Sevilla-Lara}{lsevilla@ed.ac.uk}{}

\addinstitution{
 School of Informatics\\
 University of Edinburgh, UK
}

\runninghead{Batra et.al.}{Temporal Ordering in Procedure Segmentation}

\def\etal{\emph{et al}\bmvaOneDot}
\begin{document}

\maketitle
\vspace{-1em}
\begin{abstract}
\input{sections/abstract}
\end{abstract}
\section{Introduction}
\input{sections/introduction}
\vspace{-1.5em}
\section{Related Work}
\input{sections/related_work}
\vspace{-0.8em}
\section{Evaluation of Procedure Segmentation}
\input{sections/localization}
\vspace{-0.8em}
\section{Improved Procedure Segmentation}
\input{sections/dps}
\vspace{-0.8em}
\section{Experiments and Results}
\input{sections/datasets}
\input{sections/results}
\vspace{-3em}
\section{Conclusions}
\input{sections/conclusion}
\vspace{-2em}
\section*{Acknowledgments}
\input{sections/acknowledgment}
\vspace{2em}
\bibliography{egbib}
\end{document}

%% file: sections/abstract.tex

Understanding the steps required to perform a task is an important skill for AI systems. 
Learning these steps from instructional videos involves two subproblems: (i) identifying the temporal boundary of sequentially occurring segments and (ii) summarizing 
these steps in natural language. 
We refer to this task as Procedure Segmentation and Summarization (PSS). In this paper, we take a closer look at PSS and propose three fundamental improvements over current methods. The segmentation task 
is critical, as generating a correct summary requires each step of the procedure to be correctly identified.
However, current segmentation metrics often overestimate the segmentation quality because they do not consider the temporal order of segments.
In our first contribution, we propose a new segmentation metric that takes into account the order of segments, giving a more reliable measure of the accuracy of a given predicted segmentation.
Current PSS methods are typically trained by proposing segments, matching them with the ground truth and computing a loss. 
However, much like segmentation metrics, existing matching algorithms do not consider the temporal order of the mapping between candidate segments and the ground truth. In our second contribution, we propose a matching algorithm that constrains the temporal order of segment mapping, and is also differentiable. 
Lastly, we introduce multi-modal feature training for PSS, which further improves segmentation. We evaluate our approach on two instructional video datasets (YouCook2 and Tasty) and observe an improvement over the state-of-the-art of $\sim7\%$ and $\sim2.5\%$ for procedure segmentation and summarization, respectively.
\begin{figure}[h]
    \centering
    \includegraphics[width=0.95\linewidth]{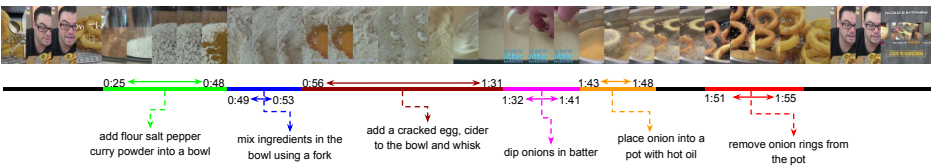}
\caption{
Procedure Segmentation and Summarization: An example from
the \textbf{YouCook2} dataset \cite{ZhXuCoCVPR18} for the procedure  \textit{Making Onion Rings}. Each procedure segment has annotated time boundaries and is summarized by an English sentence. The black segments refer to background.}
\label{fig:overview_1}
\end{figure}
\begin{figure*}
    \centering
    \includegraphics[width=0.95\linewidth]{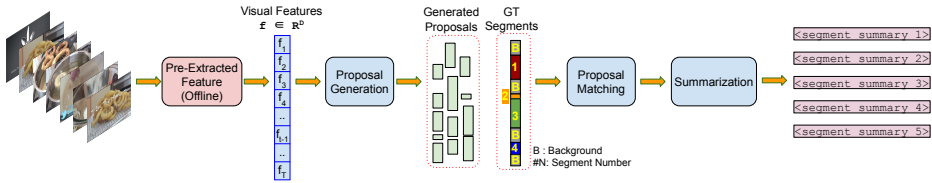}
\caption{A single stage pipeline for Procedure Segmentation and Summarization. Similar to Dense Video Captioning~\cite{krishna2017dense} with the additional constraint of including the temporal order of segments during optimization. In training, we first generate $N$ proposals with a confidence score from pre-extracted video features. Then we find sequentially ordered matched pairs between the ground truth and generated proposals to compute the loss. Finally, the matched proposals are utilized to summarize the segments.}
\label{fig:overview}
\end{figure*}

%% file: sections/introduction.tex
Temporal information is essential for understanding the evolution of objects, actions, and states in the world, and is fundamental to many vision tasks, including video summarization~\cite{Papalampidi_Keller_Lapata_2021,zhang2016video}, description~\cite{yao2015describing}, story understanding~\cite{han2019contextualized} or timeline construction~\cite{zhou2020temporal}. 
Among the possible types of temporal structure in a video, this paper addresses the problem of identifying a series of steps that occur in a particular temporal order. Such a structure is fundamental to goal-oriented tasks or procedures, such as preparing a recipe or repairing an appliance.  
Each event in the procedure is referred to as a \textit{step} or \textit{segment}~\cite{ZhXuCoCVPR18}. Together they form a complex task, like 
``making onion rings" in Fig.~\ref{fig:overview_1}. 
The example shows that the steps follow the arrow of time, i.e., segments need to occur in a particular order, and they do not overlap. These are two crucial temporal properties in instructional videos. 
For example, in Fig.~\ref{fig:overview_1}, one cannot dip onions into the batter without preparing the batter first. 


Zhou \etal~\cite{ZhXuCoCVPR18} addressed the problem of step segmentation in the context of procedures by introducing the \textit{YouCook2} annotated dataset of untrimmed instructional videos. Recent work~\cite{zhou2018end, shi2019dense} has also addressed the problem of segmentation and summarization together in instructional videos, proposing methods that target a closely related task, \textit{Dense Video Captioning} (DVC)~\cite{krishna2017dense}.  
DVC assumes segments can overlap, and there can be one or two actions per caption. By contrast, in instructional videos, segments cannot overlap, and there can be multiple small actions in a single segment (e.g., ``add a cracked egg, cider to the bowl and whisk''). Therefore the task can be interpreted as \textit{summarization}, where the objective is to generate a sentence to describe a larger amount of visual content, potentially containing multiple small actions. In this paper we refer to the task of identifying the temporal boundaries of non-overlapping steps and producing a textual summary of each step as \textit{Procedure Segmentation and Summarization} (PSS).

Previous work in PSS~\cite{ZhXuCoCVPR18, shi2019dense, wang2021end} often uses a two-stage training pipeline as in DVC~\cite{krishna2017dense}, first generating segment proposals and then describing the content of the segments (see Fig.~\ref{fig:overview}). The procedure segmentation and its evaluation metrics are critical, as summarization requires the step to be accurately identified first.
Zhou \etal~\cite{ZhXuCoCVPR18} propose to use mean-IoU 
and mean-Jaccard 
as metrics to evaluate the segmentation performance. 
These metrics do not consider temporal ordering nor enforce a one-to-one mapping between the ground truth and predicted segments. Specifically, these metrics find the proposed segment with highest overlap for each ground truth segment, and compute the average overlap. This approach ignores the crucial property of instructional videos, i.e., the annotated segments are non-overlapping and have sequential structure. 
To correctly evaluate a given segmentation, we need a metric that takes into account the ordering of segments and optimally finds matched pairs between ground truth and predicted segments. In this paper, we propose SODA-D (F-measure), a dynamic programming-based metric that finds the optimal match 
by considering their temporal order and constraint the match to be non-overlapping. This metric is inspired by SODA-C (F-measure)~\cite{fujita2020soda}, a metric for captions where ordering also matters. 
The proposed evaluation metric gives a low score when the number of generated segments is higher or lower than the number of ground truth segments. We show that current state-of-the-art methods perform worse than trivial uniform baselines under this new, more realistic evaluation framework.

Recently, Wang \etal~\cite{wang2021end} proposed PDVC, an end-to-end training method for DVC, which they also test on YouCook2 dataset. The authors use the Hungarian matching algorithm to find 1-to-1 mappings between the ground truth and generated proposals in order to optimize the procedure segmentation.
However, this matching algorithm does not consider the temporal order to find 1-to-1 mappings and can lead to inaccurate procedure segmentation. As an alternative, we propose to use a differentiable version of SODA-D as our matching algorithm to compute the loss for segmentation. This novel matching algorithm obtains state-of-the-art performance across the two datasets we test (YouCook2~\cite{ZhXuCoCVPR18} and Tasty~\cite{sener2019zero}), both under the existing metrics (mIoU and mJaccard) and the proposed SODA-D. 

Additionally, existing techniques utilize models pre-trained with a single modality. However, the current task (PSS) is inherently multi-modal. 
We propose to utilize a multi-modal features to train the PSS pipeline to incorporate both visual and textual information. 
We observe that this simple change achieves over 5\% improvement in segmentation performance, which in turn leads to improved summarization results.
%
Overall, our enhanced version of PDVC achieves a new state of the art on the YouCook2~\cite{ZhXuCoCVPR18} and Tasty~\cite{sener2019zero} datasets.
\paragraph{Contributions}
\begin{itemize}
\vspace{-0.3em}
    \itemsep0em 
    \item We show that current evaluation metrics (mean-IoU~\cite{ZhXuCoCVPR18}) for procedure segmentation misrepresent the quality of a given segmentation as they ignore the sequential order of segments. We propose to use SODA-D (F-measure)~\cite{fujita2020soda}, a dynamic programming  metric, that considers the segments' order. We show that it accurately evaluates procedure segmentation and penalizes over- and under-segmentation.
    \item Based on this evaluation metric, we propose to use its differentiable version as a loss to optimize the existing PDVC~\cite{wang2021end} model, exploiting the non-overlapping nature of procedure segments. 
    \item We explore different video feature representations and demonstrate that multi-modal features are significantly better for PSS.
    \item We apply our approach to two instructional video datasets (YouCook2~\cite{zhou2018end} and Tasty~\cite{sener2019zero}) and achieve a new state-of-the-art for PSS.
\end{itemize}

%% file: sections/related_work.tex
\paragraph{Dense Video Captioning.} Krishna \etal~\cite{krishna2017dense} proposed the dense video captioning problem in open domain videos. It is a multitask problem which focuses on performing segmentation of multiple generic events happening in the short videos and describing segmented events using sentences. Although the task that we address here (PSS) is similar to DVC, there are significant differences: (i) events in instructional videos are part of a  procedure, unlike generic events, (ii) the annotated events in open domain videos are often overlapping, while segments in instructional videos are non-overlapping and causal in nature, and (iii) the DVC task requires a description of the visual content of the event, while the PSS task requires summarizing the part of the procedure performed in a single segment (or step), which may consist of multiple actions. In contrast to prior work, we perform segmentation and summarization of procedures in instructional videos by exploiting the above key differences.

Both segmentation and summarization  hinge on pre-extracted visual features. Hence, researchers have tried to enrich the representation in different ways, e.g., using context modeling~\cite{wang2018bidirectional}, or multi-modal feature fusion~\cite{BMT_Iashin_2020}. 
The majority of existing work uses features learned from action recognition datasets. Instead, we use models pretrained with multi-modal objectives. Recently, Wang \etal~\cite{wang2021end} introduced a single-stage pipeline for DVC and utilize the Hungarian matching algorithm to connect the two stages. However, the matching algorithm does not consider the temporal order of segments in instructional videos. In this work, we propose to constrain the matching algorithm to incorporate temporal order.

Recently, Fujita \etal~\cite{fujita2020soda} challenged the established evaluation framework for DVC captions, which ignored temporal order of captions and proposed a new caption evaluation framework called SODA-C. In this evaluation metric the authors focus only on temporal captions and the corresponding segments, which can overlap. As the segments can overlap in ActivityNet~\cite{krishna2017dense} dataset, hence they did not consider event segmentation metrics.
Taking inspiration from their work, we go beyond the current evaluation framework for procedure segmentation in instructional videos~\cite{ZhXuCoCVPR18} and propose SODA-D.
\paragraph{Instructional Video Understanding.} 
Instructional videos are a good choice for learning procedural knowledge~\cite{lin2022learning} and procedure planning~\cite{chang2020procedure}. To understand instructional videos, researchers proposed various datasets ~\cite{tang2019coin, zhukov2019cross, kuehne2014language, elhamifar2019unsupervised, ZhXuCoCVPR18, sener2019zero}. Annotations in most datasets consist of a fixed label space for actions or steps in instructional videos.
Fixed labels make it easy to match ground-truth and predictions, leading to a better metric. 
However, our work focuses on instructional videos where procedure segments are summarized using natural language sentences as opposed to fixed labels. Consequently, these datasets~\cite{ZhXuCoCVPR18, sener2019zero} are more realistic, but also more challenging to work with, and procedure segmentation evaluation is more difficult. In this work, we focus on the metrics to evaluate procedure segmentation and propose techniques to improve segmentation performance.

%% file: sections/localization.tex
Current evaluation metrics~\cite{ZhXuCoCVPR18, wang2021end,mun2019streamlined} for procedure segmentation can be categorized into two groups. One group of metrics is based on proposal detection at different IoU thresholds, which includes precision, recall and F1 with limited proposals (e.g., top-10 proposals). Another group measures the overlap between ground truth and proposal segments, which includes Jaccard and mean-IoU. Let $\mathcal{G}$ and $\mathcal{P}$ be the set of ground truth and top-N predicted proposals, and $g$ a ground truth segment and $p$ a predicted segment.

\begin{figure*}
\begin{center}
\includegraphics[width=0.9\linewidth]{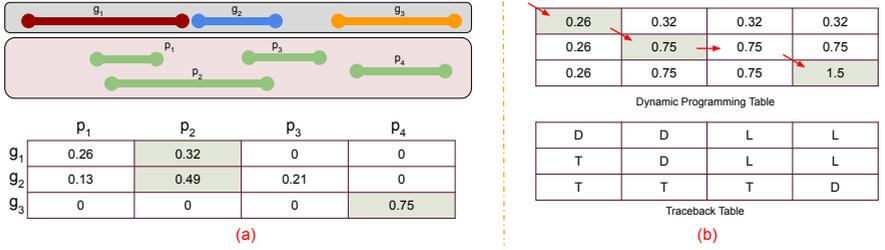}
\end{center}
  \caption{Issues with evaluation metrics for procedure segmentation. (a) \textbf{Top}: A hypothetical example of ground truth ($g_i$) and generated proposal ($p_i$) segments. \textbf{Bottom}: Pair-wise IoU between each ground truth and generated proposal. (b) \textbf{Top}: Filled dynamic programming~\cite{fujita2020soda} table corresponding the IoU cost matrix. \textbf{Bottom}: Traceback matrix, where L, T, D signify left, top and diagonal path.}
\label{fig:challenge}
\end{figure*}

\paragraph{Issues with Current Metrics:} 
We argue that the evaluation metric for segmentation utilized in most prior work~\cite{mun2019streamlined,wang2021end} is inappropriate for procedure segmentation, as it only measures the quality of generated proposals.
To compute that metric, we first find the set of predicted segments ($P_{g,\tau}$) for each ground truth whose IoU exceeds a specific threshold ($\tau$). Similarly, we find the set of ground truth segments ($G_{p,\tau}$). Then we compute precision and recall using Eqn.~\ref{eq2}.
\begin{equation}\label{eq1}
\begin{aligned}[c]
P_{g,\tau} &= \{p\in\mathcal{P} | IoU(g,p) > \tau\}
\end{aligned}
\qquad \qquad
\begin{aligned}[c]
G_{p,\tau} &= \{g\in\mathcal{G} | IoU(g,p) > \tau\} 
\end{aligned}
\end{equation}
\vspace{-1.2em}
\begin{equation}\label{eq2}
\begin{aligned}[c]
Precision &= \frac{|\bigcup_{g\in\mathcal{G}} P_{g,\tau}|}{|\mathcal{P}|}
\end{aligned}
\qquad \qquad
\begin{aligned}[c]
Recall &= \frac{|\bigcup_{p\in\mathcal{P}} G_{p,\tau}|}{|\mathcal{G}|}
\end{aligned}
\end{equation}
%
This metric fails to find a one-to-one mapping between generated and ground truth segments and has the tendency to map one proposal with multiple ground truth segments, leading to inaccurate evaluation results. 
Consider the hypothetical example in Fig.~\ref{fig:challenge} with three ground truth segments and four predicted proposals. Using Eqn.~\ref{eq1} and Eqn.~\ref{eq2}, the precision and recall would be 0.5 and 1.0 for $\tau=0.3$ (see Eqn.~\ref{eq3}). Note that proposal $p_2$ is mapped to $g_1$ and $g_2$, which is inaccurate and gives incorrect detection results. The correct recall should be 0.66 ($\frac{|\{g_2, g_3\}|}{3} = \frac{2}{3}$), as the ground truth segment $g_1$ is not matched with any proposal. 
\begin{equation}\label{eq3}
\begin{aligned}[c]
\text{For} \; \tau=0.3; \qquad Precision &= \frac{|\{p_2, p_4\}|}{4} = 0.5 \\
\end{aligned}
\qquad
\begin{aligned}[c]
Recall &= \frac{|\{g_1, g_2, g_3\}|}{3} = 1.0
\end{aligned}
\end{equation}
Secondly, the overlap metrics~\cite{ZhXuCoCVPR18} compute the mean-IoU and mean-Jaccard between ground truth and predicted top-N segments. These metrics do not consider the temporal order and one-to-one mapping between ground truth and predicted segments. Specifically, they iteratively find the best matching segment with highest overlap for each ground truth segment and compute the average (see Eqn.~\ref{eq4}). The inherent property of instructional videos, i.e., that annotated procedure segments are non-overlapping and have sequential structure, is ignored.
%
\begin{align}\label{eq4}
    mIoU &= \frac{\sum_{g\in\mathcal{G}} max({IoU(g,p) | p\in\mathcal{P}})}{|\mathcal{G}|}
\end{align}
For example, $g_1,g_2$ achieve maximum overlap with $p_2$ and are selected to compute the average IoU. This leads to an overestimate in the current metric, i.e., 0.52 ($\frac{0.32 + 0.49 + 0.75}{3}$) in contrast to 0.5 ($\frac{0.26 + 0.49 + 0.75}{3}$), as we have obtained only two proposals ($p_2, p_4$) to compare with three ground truth segments. Hence, we need a metric which penalizes selecting a single predicted proposal for multiple ground truth segment and vice versa.

\paragraph{Prerequisite Soda-C Metric:} SODA-C (F-measure) is proposed by Fujita \etal~\cite{fujita2020soda} to evaluate the coherent story of sequential captions which are generated by Dense Video Captioning models, e.g., PDVC~\cite{wang2021end}. 
SODA-C finds the optimal sequence of generated captions by maximizing the joint score of the proposal IoU and the corresponding caption METEOR score. The metric utilizes a dynamic programming based algorithm to find the optimal match between generated and ground truth captions considering the temporal ordering of captions. The SODA-C score is penalized if the number of generated captions is different from the ground truth and the order of captions is incorrect (e.g.,~the two ground truth captions are swapped in prediction). Fujita \etal~\cite{fujita2020soda} work only focus on the evaluation of temporal captions and the corresponding overlapping segments (e.g. ActivityNet dataset~\cite{krishna2017dense}), in contrast our task consists of non-overlapping annotated segments. Hence they did not consider segmentation metrics. Inspired by the characteristics of the SODA-C metric and the issues in procedure segmentation we discussed, i.e., the need of temporal order in procedure segments, we propose a modified version of SODA-C for the evaluation of procedure segmentation.
\paragraph{Sequential Matching-based Evaluation:}
As explained previously, the current evaluation framework for DVC is inaccurate for the procedure segmentation task due to its inability to find 1-to-1 mappings between ground truth and generated proposals. Inspired by 
SODA-C~\cite{fujita2020soda}, a dynamic programming-based algorithm to find matched pairs between the generated captions and ground truth captions, we propose the SODA-D metric\footnote{SODA-D is not described in the original paper~\cite{fujita2020soda}. However, their code contains and utilizes SODA-C.} with a different cost matrix and objective. The proposed metric finds 1-to-1 mapping and computes the optimal mean-IoU for procedure segmentation (in contrast to finding optimal captions). The significant property of the algorithm is that it considers the temporal order of segments to find the matching, and therefore is suitable for the non-overlapping segments in instructional videos. We compute the IoU cost matrix between every ground truth and predicted proposal and apply the recurrence equation (Eqn.~\ref{eq5}) to fill the dynamic programming table. Finally, we trace back the diagonal elements to find the matched pairs and compute the mean-IoU. We refer the reader to SODA~\cite{fujita2020soda} for more information and pseudo code.
\begin{equation}\label{eq5}
\begin{aligned}[c]
C_{i,j} &= IoU(g_i, p_j)\\
\end{aligned}
\qquad \qquad
\begin{aligned}[c]
S[i][j] &= max \begin{cases}
      S[i-1][j] \\
      S[i-1][j-1] + C_{i,j}\\
      S[i][j-1]
    \end{cases} 
\end{aligned}
\end{equation}
Here, $C_{i,j}$ represents the $(i,j)$ element of the cost matrix, as shown in Fig.~\ref{fig:challenge}(a) and $S[i][j]$ represents the dynamic programming table to store the maximum score for optimal matching. 

Following the earlier hypothetical example, we apply the recurrence equation and obtain the dynamic programming table as shown in Fig.~\ref{fig:challenge}(b). Tracing the diagonal elements (shown as $D$ in Fig.~\ref{fig:challenge}(b), bottom) we obtain the optimal one-to-one mapping between ground truth and predicted proposals, leading us to compute the accurate mean-IoU. Similarly, we can compute the precision/recall/F1 score that penalizes inappropriate proposals using Eqn.~\ref{eq6}:
\begin{equation}\label{eq6}
\begin{aligned}[c]
Precision &= \frac{\sum_{g\in\mathcal{G}} IoU(g,a(p))}{|\mathcal{P}|}
\end{aligned}
\qquad \qquad
\begin{aligned}[c]
Recall &= \frac{\sum_{g\in\mathcal{G}} IoU(g,a(p))}{|\mathcal{G}|}
\end{aligned}
\end{equation}
where $a(p)$ corresponds to matched proposal for the ground truth segment ($g$) obtained via tracing through the dynamic programming table ($S$).

%% file: sections/dps.tex
\begin{figure*}
\begin{center}
\includegraphics[width=0.9\linewidth]{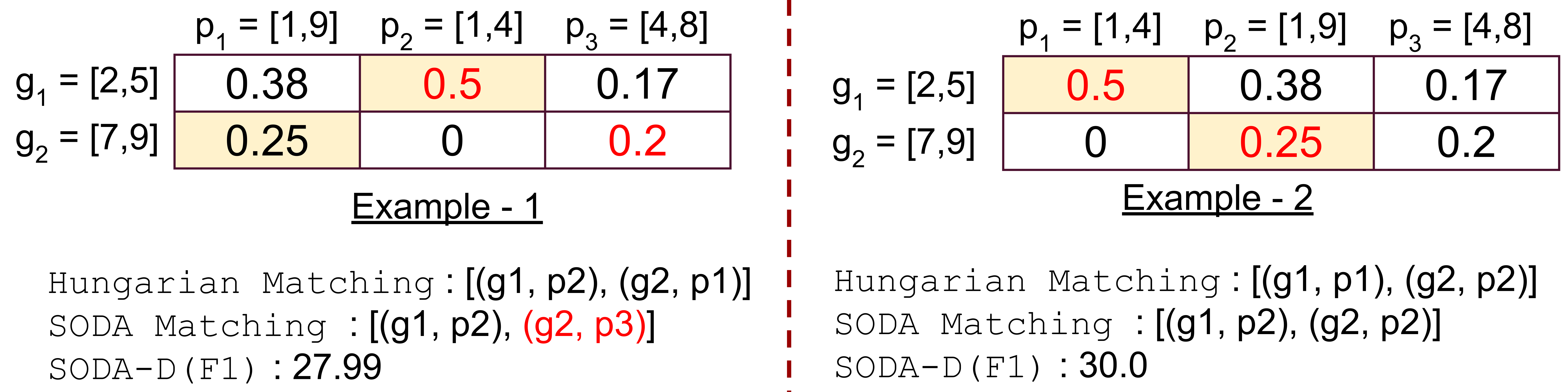}
\end{center}
  \caption{Hypothetical Example~1 shows that the Hungarian algorithm does not consider the temporal order of segments. Additional hypothetical Example~2 with swapped proposals shows that the proposal ordering affects the SODA-D score.}
\label{fig:hungchallenge}
\end{figure*}
\paragraph{Video Feature Encoding:}
Prior work uses a pre-trained action recognition network (e.g., I3D, TSN) to pre-compute frame or clip level features. Pre-trained networks focus only on the visual modality of the action recognition task. However, our task is multi-modal, i.e., we need temporal and semantic features of videos to perform segmentation and summarization of the segments appropriately. Based on this intuition, we hypothesize that 
if we extract features offline from an architecture pre-trained with a multi-modal objective for video-text representation learning  will be an improvement over video-only features e.g., the S3D~\cite{miech19endtoend} architecture.
Moreover, the multi-modal video features also encode semantic and conceptual information from the text. They can be utilized for instructional videos with missing transcripts, e.g., Tasty~\cite{sener2019zero}. We therefore explore features from the S3D~\cite{miech19endtoend} network pre-trained on instructional videos with a multi-modal objective.
\paragraph{Optimal Sequential Proposal Matching:} 
Recently, Wang \etal~\cite{wang2021end} proposed a single-stage pipeline (PDVC) based on the DETR~\cite{carion2020end} architecture and connect the two stages with a set-based matching loss to perform dense video captioning in parallel. 
The authors (i) extract the video features offline, (ii) generate proposals with confidence scores using a transformer encoder-decoder architecture, (iii) find optimal 1-to-1 mapping between GT and generated proposals via the Hungarian algorithm, and (iv) generate captions for matched proposals. The decoder outputs a segment counter (number of segments in the video) and picks the top-$N$ segments from sorted proposals based on a confidence score. The set-based loss in PDVC~\cite{wang2021end} utilizes the Hungarian algorithm to find a match between the ground truth and predicted segments in order to compute the regression loss on the temporal boundaries. 
However, the direct application of the Hungarian algorithm in PSS makes the matching loss inappropriate, as it does not consider the temporal order to find the matched pairs and can lead to inaccurate procedure segmentation. Example~1 in Fig.~\ref{fig:hungchallenge} shows that the Hungarian algorithm chooses temporally incorrect matches, i.e., $[(g_1,p_2),(g_2,p_1)]$ in contrast to temporally correct matches, i.e., $[(g_1,p_2),(g_2,p_3)]$. The temporally correct matching helps in the optimization of boundary points regression, as the boundary points of $p_3$ proposals are closer to $g_2$. Example~2 shows that the sequential order changes the SODA-D score.

To tackle this issue, we propose a differentiable version of the matching algorithm utilized in our evaluation metric (SODA-D) and modify PDVC~\cite{wang2021end} by replacing the Hungarian algorithm with our differentiable matching algorithm. It enforces 1-to-1 mappings and follows the temporal order in the ground truth and predicted segments. We convert the IoU maximization problem of the matching algorithm (Eqn.~\ref{eq5}) into minimization by negating the cost ($C_{i,j} = -IoU(g_i,p_j)$) and replacing $max$ with $min$ in the recurrence equation. 
Further, we replace the non-differentiable hard $min$ operator with the $smoothMin$ operator~\cite{hadji2021representation}. This makes the whole PSS training pipeline differentiable. Due to 1-to-1 mapping in the temporal order, the proposed differentiable SODA algorithm penalizes overlapping segments.

%% file: sections/datasets.tex
\vspace{-0.3em}
\subsection{Datasets}
We perform our experiments on two instructional video datasets, YouCook2~\cite{ZhXuCoCVPR18} and Tasty~\cite{sener2019zero}. Both datasets contain cooking videos, which are fully annotated with  temporal boundaries and English sentences. 
\paragraph{YouCook2~\cite{ZhXuCoCVPR18}:} The dataset contains videos of 89 different cooking procedural tasks. The official split has 1333/457 videos in the training and validation sets. At the time of writing, we can download 1203 videos from the official set and we create our own splits. The training set consists of 840, the validation set of 183, and the test set of 180 videos. The average cooking procedure has a duration of 320s, and each video has 7.7 annotated segments with associated sentences on average. All videos are unconstrained and in third-person viewpoint.
\paragraph{Tasty~\cite{sener2019zero}:} This dataset has 2511 unique recipes belonging to 185 tasks, such as making cakes, pies, and soups. The Tasty videos are captured with a fixed overhead camera and focus entirely on the dish preparation. The videos are short, average duration is 54s and contain an average of nine instructions per recipe. We utilize the official split for training/validation/testing of 2998/399/400 videos.
\vspace{-0.9em}
\subsection{Implementation Details}
\paragraph{Training Details:} 
We follow PDVC~\cite{wang2021end} to extract clip-level visual features for both datasets and use the same hyper-parameters to train the PSS pipeline.
Similar to~\cite{wang2021end}, we pre-compute the clip-level visual features for instructional videos for both datasets at a fixed framerate of 16 fps. The visual features are resized to a fixed temporal size of 200/150 for YouCook2/Tasty. Following~\cite{wang2021end}, we train the network with a batch size of one for 30 epochs and use the Adam optimizer with an initial learning rate of 5e-5. We use the same hyper-parameters as in PDVC\footnote{PDVC Code available at \url{https://github.com/ttengwang/PDVC}}~\cite{wang2021end} and replace the matching algorithm with ours. To implement the matching algorithm, we utilize the dynamic programming function from SODA\footnote{SODA Matcher; dynamic programming function available at: \url{https://github.com/fujiso/SODA}}~\cite{fujita2020soda}. Lastly, we utilize SODA-C~\cite{fujita2020soda} and standard language generation metrics to evaluate procedure summarization. 
\paragraph{Baselines:} We compare our evaluation framework and proposed matching algorithm with three uniform baselines. (i) Uniform (Avg \#Segments): we compute the average number of segments ($n$) per video in the dataset and divide each video into $n$ equal segments. 
The average number of segments is 8 and 10 for YouCook2 and Tasty, respectively. (ii) Uniform (Avg \#Duration):  we compute the average duration of segments $d$ per video in the dataset and divide the video into equal segments of length $d$. 
The average duration of segments is 19.6 and 6 seconds for YouCook2 and Tasty, respectively. (iii) Uniform (GT):  We divide the video into equal segments assuming we know the ground-truth segment count for each video. Additionally, we compare to the performance of two competitive methods, ProcNet~\cite{zhukov2019cross} and the original version of PDVC~\cite{wang2021end}. We re-implement ProcNet\footnote{ProcNet Code in LuaTorch available at \url{https://github.com/LuoweiZhou/ProcNets-YouCook2}}~\cite{zhukov2019cross} in PyTorch. We train both models on our training set with four runs and report the result on the test set.

%% file: sections/results.tex
\begin{table*}
\scriptsize
\renewcommand{\tabcolsep}{0.1cm}
\centering
    \begin{tabular}{lccccc}
        \toprule[1.5pt]
\specialcell{\bf Method} & \multicolumn{2}{c}{\specialcell{\bf Existing}} & \multicolumn{3}{c}{\specialcell{\bf Proposed}}\\
 \cmidrule(lr){2-3}\cmidrule(lr){4-6}
 & mIoU & mJaccard & Precision & Recall & SODA-D (F1)\\ \midrule
Uniform (Avg \#Segments) & 35.18 & 40.79 & 29.01 & 31.67 & 29.46 \\
Uniform (Avg \#Duration) & 43.43 & 61.87 & 22.50 & 43.77 & 28.53 \\
Uniform (GT) & 34.18 & 38.26 & 30.47 & 30.47 & 30.47 \\
ProcNet ~\cite{ZhXuCoCVPR18} & 32.30 & 46.32 & 26.72 & 30.98 & 27.89 \\
PDVC ~\cite{wang2021end} & 33.54 & 42.61 & 27.44 & 31.39 & 28.35 \\
Ours & \textbf{41.38} & \textbf{52.63} & \textbf{36.01} & \textbf{39.67} & \textbf{36.80} \\
 \bottomrule[1.5pt] \\
    \end{tabular}
    \caption{Comparison of the existing evaluation metric and the proposed SODA-D metric for procedure segmentation. The average results of models with four different seeds.}
    \label{table:prior_eval}
\end{table*}
\vspace{-1em}
\subsection{Ablation Study}
This section compares the existing overlap-based metric with the proposed SODA-D score for procedure segmentation evaluation. We also investigate the effects of multiple design choices and discuss the results on the YouCook2~\cite{ZhXuCoCVPR18} dataset.
\paragraph{Optimal Matching-based Evaluation:} We compare the different metrics in Table~\ref{table:prior_eval}. We compare Recall and mIoU, as they are equivalent (Eqns.~\ref{eq3} and~\ref{eq5}), i.e., mIoU focuses on maximum overlap without order in contrast to the proposed metric. The uniform baseline with known ground-truth and average segments shows that the overlap-based metric overestimates the method's performance by 3 to 4\%. The current overlap-based metric fails to penalize over-segmentation, e.g., the uniform baseline with average duration has similar mIoU and Recall. However, precision is low due to the large number of proposals leading to a lower SODA-D
score. We found that recent techniques developed for PSS perform worse or the same as the uniform baseline on the proposed evaluation metric. Our approach outperforms the uniform baseline by a significant margin of~6\%.
\paragraph{Visual Features:} It is not feasible to learn features from long untrimmed instructional videos 
due to high GPU memory demand. The feasible alternative is pre-extracting the video features from models pre-trained on clip-level videos. Following prior work~\cite{ZhXuCoCVPR18, wang2021end}, we explore different frame level (TSN~\cite{8454294}) and clip level features (R3D~\cite{hara3dcnns}, I3D~\cite{carreira2017quo}, S3D~\cite{miech19endtoend}). We find that features extracted from a multi-modal model perform better than any alternative. 
The results show that multi-modal features (S3D~\cite{miech19endtoend}) improve procedure segmentation and summarization  by a significant margin of $\sim5\%$ and $\sim2.5\%$, respectively (row 4 in Table~\ref{table:ablation}).
\paragraph{SODA Matching:} We hypothesize that temporal order information is required in the matching algorithm to generate non-overlapping proposals with high confidence. To test our hypothesis, we experiment with the non-differentiable SODA matching algorithm replacing the Hungarian algorithm and show the results in Table~\ref{table:ablation}. We observe that the proposed matching algorithm improves the results by $\sim2.5\%$ for procedure segmentation, confirming our hypothesis. The differentiable version (SoftSODA) marginally improves the performance for procedure segmentation further. Lastly, the results show that better procedure segmentation also improves summarization performance.
\begin{table*}
\scriptsize
\renewcommand{\tabcolsep}{0.2cm}
\centering
    \begin{tabular}{lccc}
        \toprule[1.5pt]
\specialcell{\bf Features} & \specialcell{\bf Matcher} & \specialcell{\scriptsize SODA-D} & \specialcell{\scriptsize SODA-C}\\ \midrule
TSN~\cite{8454294} & Hungarian & 27.44 ± 1.56 & 3.80 ± 0.26 \\
ResNeXt-3D (R3D)~\cite{hara3dcnns} & Hungarian & 28.36 ± 0.28 & 4.11 ± 0.05 \\
I3D~\cite{carreira2017quo} & Hungarian & 28.23 ± 1.41 & 4.17 ± 0.28 \\
S3D~\cite{miech19endtoend} & Hungarian & 33.10 ± 0.28 & 6.13 ± 0.08 \\
S3D~\cite{miech19endtoend} & SODA & 36.32 ± 2.19 & 6.39 ± 0.51 \\
S3D~\cite{miech19endtoend} & SoftSODA & \textbf{36.80 ± 1.90} & \textbf{6.54 ± 0.44} \\
 \bottomrule[1.5pt] \\
    \end{tabular}
    \caption{Ablation study on the YouCook2~\cite{ZhXuCoCVPR18} dataset for different visual features and matching algorithm utilizing the PDVC~\cite{wang2021end} model architecture.}
    \label{table:ablation}
\end{table*}
\begin{table*}
\scriptsize
\renewcommand{\tabcolsep}{0.15cm}
\centering
    \begin{tabular}{lcccccc}
        \toprule[1.5pt]
\specialcell{\bf Method} & \multicolumn{3}{c}{\specialcell{\bf YouCook2}} & \multicolumn{3}{c}{\specialcell{\bf Tasty}}  \\
 \cmidrule(lr){2-4}\cmidrule(lr){5-7}
 & Precision & Recall & SODA-D (F1) & Precision & Recall & SODA-D(F1) \\ \midrule
Uniform (Avg \#Segments) & 29.01 & 31.67 & 29.46 & 35.11 & 42.47 & 36.82 \\
Uniform (Avg \#Duration) & 22.50 & 43.77 & 28.53 & 37.83 & 44.71 & 39.88 \\
Uniform (GT) & 30.47 & 30.47 & 30.47 & 43.35 & 43.35 & 43.35 \\
ProcNet ~\cite{ZhXuCoCVPR18}, AAAI 2018 & 26.72 ± 1.01 & 30.98 ± 1.63 & 27.89 ± 1.25 & 29.87 ± 3.07 & 44.05 ± 3.51 & 34.12 ± 1.01 \\
PDVC ~\cite{wang2021end}, ICCV 2021 & 27.44 ± 0.49 & 31.39 ± 0.72 & 28.35 ± 0.27 & 44.83 ± 0.97 & 42.07 ± 0.61 & 42.44 ± 0.66 \\
Ours & \textbf{36.01 ± 2.06} & \textbf{39.67 ± 1.79} & \textbf{36.80 ± 1.90} & \textbf{52.92 ± 0.85} & \textbf{50.18 ± 0.58} & \textbf{50.37 ± 0.63} \\
 \bottomrule[1.5pt] \\
    \end{tabular}
    \caption{Comparison of our improvements with state-of-the-art methods on the procedure segmentation task. Ours is the full model with proposed improvements, i.e., multi-modal features (S3D~\cite{miech19endtoend}) and differentiable matching (SoftSODA) algorithm considering temporal order.}
    \label{table:results_loc}
\end{table*}
\vspace{-1.2em}
\subsection{Comparison to State of the Art} We compare the effectiveness of the proposed method to the state-of-the-art models ProcNet~\cite{ZhXuCoCVPR18} and PDVC~\cite{wang2021end} (Table~\ref{table:results_loc} and Table~\ref{table:results_cap}). PDVC~\cite{wang2021end} marginally improves over ProcNet~\cite{ZhXuCoCVPR18} as the dense video captioning model is inappropriate due to the Hungarian matching algorithm that generates overlapping proposals. ProcNet~\cite{ZhXuCoCVPR18} benefits from the sequential generation of proposals using an LSTM. However, it utilizes fixed hand-crafted anchors to generate proposals, leading to poor performance for procedure segmentation. Our proposed modification of PDVC~\cite{wang2021end} improves by learning to generate non-overlapping proposals via the SODA matching algorithm. This improves the segmentation performance by $\sim7\%$ on both datasets. Moreover, with better predicted segments, the system also improves summarization (i.e., summarizing the activities in generated segments) by $\sim2.5\%$ and $\sim3\%$ for YouCook2 and Tasty, respectively. See supplementary material for qualitative comparisons.
\begin{table}
\scriptsize
\renewcommand{\tabcolsep}{0.1cm}
\centering
    \begin{tabular}{lcccccccc}
        \toprule[1.5pt]
\specialcell{\bf Method} & \multicolumn{4}{c}{\specialcell{\bf YouCook2}} & \multicolumn{4}{c}{\specialcell{\bf Tasty}}  \\
 \cmidrule(lr){2-5}\cmidrule(lr){6-9}
 & B4 & METEOR & CIDER & SODA-C & B4 & METEOR & CIDER & SODA-C \\ \midrule
\multicolumn{9}{c}{\specialcell{\bf Ground Truth Proposals}} \\
PDVC~\cite{wang2021end} & 1.46 ± 0.27 & 10.39 ± 0.14 & 52.42 ± 1.54 & 13.61 ± 0.12 & 4.21 ± 0.11 & 10.98 ± 0.07 & 72.99 ± 1.44 & 15.19 ± 0.14 \\
Ours & \textbf{3.27 ± 0.22} & \textbf{13.58 ± 0.16} & \textbf{83.84 ± 1.14} & \textbf{17.85 ± 0.07} & \textbf{6.83 ± 0.15} & \textbf{14.29 ± 0.12} & \textbf{102.81 ± 0.83} & \textbf{18.89 ± 0.10}  \\
\midrule
\multicolumn{9}{c}{\specialcell{\bf Learned Proposals}} \\
PDVC~\cite{wang2021end} & 0.61 ± 0.15 & 3.79 ± 0.04 & 19.14 ± 0.53 & 4.11 ± 0.05 & 2.63 ± 0.06 & 7.34 ± 0.05 & 35.85 ± 1.18 & 6.58 ± 0.17 \\
Ours & \textbf{0.95 ± 0.20} & \textbf{5.13 ± 0.56} & \textbf{27.20 ± 3.56} & \textbf{6.54 ± 0.44} & \textbf{3.82 ± 0.32} & \textbf{8.54 ± 0.29} & \textbf{48.58 ± 2.95} & \textbf{9.63 ± 0.21} \\
 \bottomrule[1.5pt] \\
    \end{tabular}
    \caption{Comparison of procedure segment summarization with generated and ground-truth proposals. Ours is the full model with proposed improvements.}
    \label{table:results_cap}
\end{table}

%% file: sections/conclusion.tex
\vspace{-0.8em}
In summary, this work uncovered issues in current evaluation metrics for the temporal segmentation of non-overlapping procedures in instructional videos. Current metrics do not find a one-to-one mapping between the ground truth and generated proposals. We proposed a new evaluation framework based on the optimal matching before computing the mIoU and demonstrate that the results obtained by current metrics are inaccurate. We also utilize multi-modal pre-trained models to extract video features and propose a differentiable optimal matching algorithm to compute the set loss 
in the proposal generation module. Finally, we show the effectiveness of our improvement on two instructional video datasets and achieve a considerable improvement over current methods.

%% file: sections/acknowledgment.tex
\vspace{-0.9em}
This work was supported in part by the UKRI Centre for Doctoral Training in Natural Language Processing, funded by UKRI grant EP/S022481/1 and the University of Edinburgh, School of Informatics.